\pdfoutput=1
\documentclass[letterpaper, 10pt,conference]{ieeeconf}

\IEEEoverridecommandlockouts                              

\usepackage{multicol}
\usepackage{hyperref}
\hypersetup{colorlinks,allcolors=black}

\usepackage{amsmath}
\usepackage{amssymb}
\usepackage{array}
\usepackage{algorithm}
\usepackage[noend]{algpseudocode}
\usepackage{float}
\usepackage[capitalise]{cleveref}
\usepackage[binary-units=true]{siunitx}
\usepackage[caption=false]{subfig}
\usepackage{balance}
\usepackage{soul}
\usepackage{tikz}
\usepackage{accents}
\usepackage{mathtools}
\usepackage{graphbox}
\usepackage{adjustbox}

\usepackage{mwe}
\usepackage{xcolor}
\def\checkmark{\tikz\fill[scale=0.4](0,.35) -- (.25,0) -- (1,.7) -- (.25,.15) -- cycle;}
\DeclareMathOperator*{\argmax}{argmax}
\DeclareMathOperator*{\argmin}{argmin}

\begin{document}

\title{\LARGE \bf
Compositional Scalable Object SLAM}

\author{Akash Sharma, Wei Dong, and Michael Kaess%
\thanks{The authors are with the Robotics Institute, Carnegie Mellon University, Pittsburgh, PA 15213, USA
        {\tt\small \{akashsha, weidong, kaess\}@andrew.cmu.edu}}
}





\maketitle

\begin{abstract}

 We present a fast, scalable, and accurate Simultaneous Localization and Mapping (SLAM) system that represents indoor scenes as a graph of objects. Leveraging the observation that artificial environments are structured and occupied by recognizable objects, we show that a compositional scalable object mapping formulation is amenable to a robust SLAM solution for drift-free large scale indoor reconstruction.  To achieve this, we propose a novel semantically assisted data association strategy that obtains unambiguous persistent object landmarks, and a 2.5D compositional rendering method that enables reliable frame-to-model RGB-D tracking. Consequently, we deliver an optimized online implementation that can run at near frame rate with a single graphics card, and provide a comprehensive evaluation against state of the art baselines. An open source implementation will be provided at \texttt{https://placeholder}.

\end{abstract}

\section{Introduction} \label{sec: introduction}

Autonomous robots that work in the \textit{real world} require advanced interpretation of the world, from semantic 3D reconstruction, path planning, to active interaction with the environment. These workloads require not only geometric perception including robot localization and dense scene reconstruction, but also semantic and compositional understanding of scenes.


In recent years, geometry-based SLAM has achieved high levels of performance in \textit{experimental setups} for localization tasks. Many variants of SLAM algorithms, from ORB-SLAM~\cite{mur-artal_orb-slam2_2017} to Direct Sparse Odometry (DSO)~\cite{engel2017direct}, can now run in real-time with high trajectory accuracy.
However, they are in general limited by the \textit{static-world} assumption and low-level scene representation (sparse 3D feature points), and thus cannot distill high-level information (semantic understanding) in scenes and adjust to structured enviromental changes.

On the other hand, with the progress in deep learning, near frame rate semantic perception is achievable powered by efficient Deep Neural Networks (DNNs).
Researchers have started to switch to semantic SLAM taking advantage of off-the-shelf solutions. Pioneering research includes SLAM++~\cite{salas-moreno_slam_2013}, Fusion++~\cite{fusionPP}, and MaskFusion~\cite{runz_maskfusion_2018}. These initial attempts take into consideration semantic segmentation, but typically simply attach DNN frontends to existing SLAM frameworks in an ad hoc fashion. Implementation-wise, they require high-end machines to achieve near real-time performance, or are not available to the community.

To address these problems, we propose a novel modular solution that concentrates on recognizable persistent object landmarks. In theory, we derived a compositional scalable object map for robust tracking. In implementation, we fully exploit the power of the modern GPU-based reconstruction pipeline~\cite{dong2019gpu} and object detection frameworks ~\cite{kirillov_pointrend_2020}, to design an efficient architecture for data exchange without sacrificing the ease of system configuration and build.



\begin{figure}[t!]
    \centering
    \includegraphics[width=\linewidth]{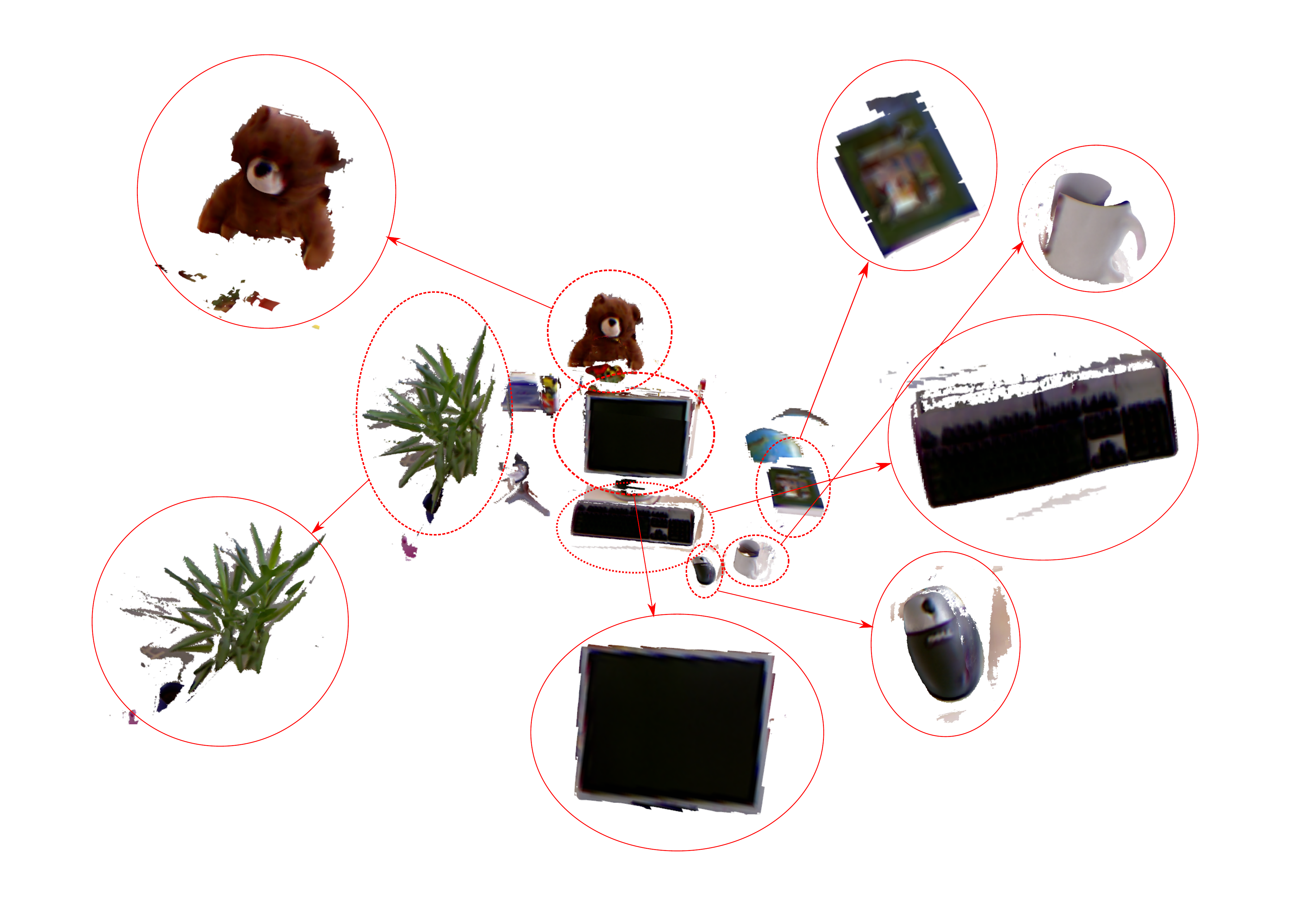}\vspace{-1cm}
    \caption{Reconstruction of \textit{fr2\_xyz} sequence from \emph{tum rgbd} dataset. Our pipeline can reconstruct both camera trajectory and object models in the scene.}
    \label{fig:objectsandscene}
\end{figure}

Our main contributions in this paper are:

\begin{enumerate}
    \item A compositional volumetric rendering method that selects objects of interest and reduces memory footprint;
    \item A hybrid object association method joining geometric and semantic cues, enabling drift-free tracking without an explicit relocalization module;
    \item A scalable, modular, and easy-to-use open source system that runs nearly realtime.
\end{enumerate}

 \begin{figure*}[ht!]
 	\centering
    \includegraphics[width=0.95\linewidth]{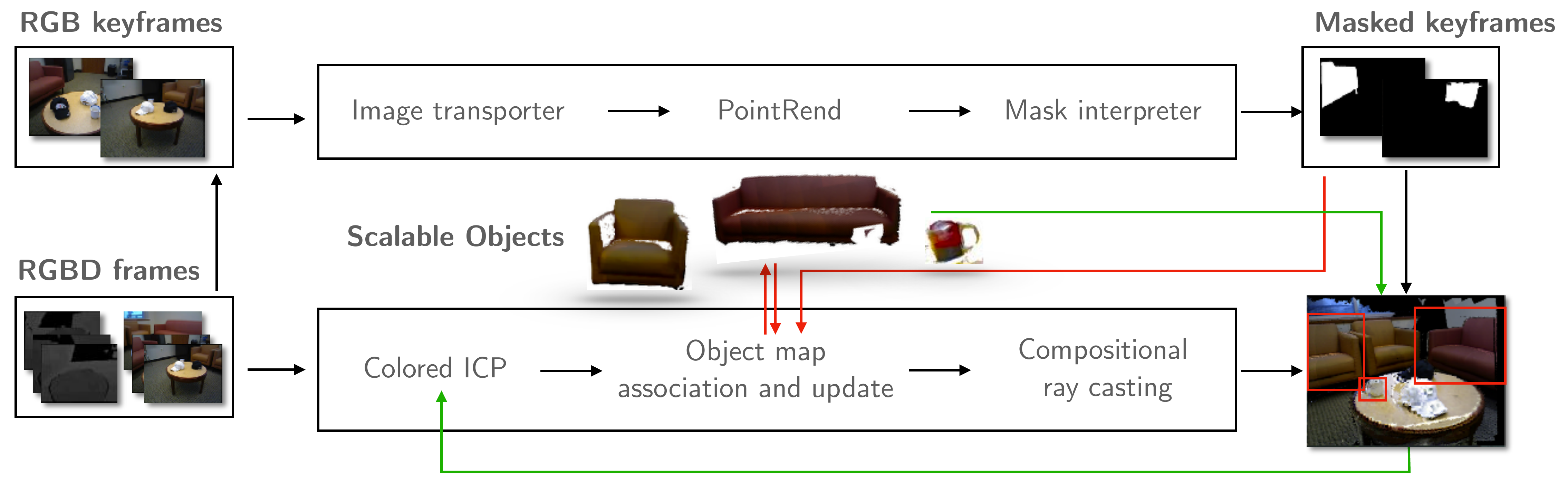}
    \caption{\label{fig:overview} System overview: Top shows the deep object segmentation pipeline that runs asynchronously, Masked keyframes from the segmentation pipeline are used in Data association and map update (shown with red lines). Bottom shows the major stages of the reconstruction system, specifically object models are used in tracking via compositional raycasting (shown with green lines).}
    \vspace*{-1em}
 \end{figure*}



\section{Related Work} \label{sec: related_works}

In this section we review relevant works in two aspects: classical geometry-based SLAM, and the application of deep semantic object detection in SLAM.

\subsection{Geometry-based SLAM}
\textbf{Problem formulation and pose optimization.} 
Modern geometry-based SLAM systems can be generally classified into \textit{feature-based} and \textit{direct} methods.
Feature-based SLAM systems ~\cite{mur-artal_orb-slam2_2017, klein2007parallel} usually maintain a collection of sparse 3D \textit{point landmarks} corresponding to hand-crafted feature \textit{keypoints} detected in 2D images. In order to correct accumulated \textit{pose} error, \textit{i.e.,} drift, these methods resort to bundle adjustment \cite{triggs1999bundle} that jointly minimizes reprojection error between \textit{landmarks} and 2D \textit{keypoints} via \textit{pose} optimization. While being accurate in estimating trajectory, these approaches only come with sparse 3D maps that are less interpretable for visualization and recognition.
Direct SLAM~\cite{engel2014lsd, engel2017direct} on the other hand, relies on pixel-wise projective data association between frames for odometry. Given relative poses between certain keyframes, pose graphs~\cite{dellaert_factor_2017} are formulated and optimized to obtain globally consistent poses without landmark constraints. 

Our approach can be regarded as a bridge of the two approaches. We replace point landmarks with objects in feature-based SLAM. As a result, since pose constraints attached to objects can naturally replace reprojection error, we may directly convert such a landmark-pose constraint optimization to pose graph optimization (PGO).

\textbf{Map representation}.
For \textit{dense} scene reconstruction, 
Truncated Signed Distance Function (TSDF) volumetric~\cite{curless1996volumetric} representation has been adopted and improved in several \textit{direct SLAM} frameworks. KinectFusion~\cite{kinectfusion} introduced a plain $512^3$ grid for small scenes and objects. VoxelHashing~\cite{niesner_real-time_2013} designed spatial hashing to scale this data structure to larger scenes. Similar implementations are available in CPU/GPU in the modern Open3D framework~\cite{zhou_open3d_2018, dong2019gpu}, and we adopt this representation due to its ease of use.

\subsection{Object instance segmentation and object-based SLAM}
\textbf{Object instance segmentation.} 
In recent years, \textit{Region proposal} based Convolutional Neural Networks (R-CNN) \cite{ren2015faster} have established themselves as de-facto standards for object \textit{instance segmentation} from images. Amongst the literature, \textit{Mask-RCNN}~\cite{he2017mask} and \textit{PointRend}~\cite{kirillov_pointrend_2020} are the best off-the-shelf solutions. 
In this work, we use \textit{PointRend} \cite{kirillov_pointrend_2020}, which shows significant improvement over \cite{he2017mask} by reformulating the mask generation as a rendering problem. In essence, this formulation is consistent with our tracking via compositional rendering module.

\textbf{Object-based SLAM.} Applying aforementioned DNNs on 2D images, several works for RGB-D and monocular SLAM have attempted to incorporate object instance detection. CubeSLAM~\cite{yang2019cubeslam} and QuadricSLAM~\cite{quadric_slam} fit cuboids and quadrics, respectively, to detected objects to generate parameterized object landmarks. While improving the localization accuracy compared to baselines, these methods fail to densely map objects. \textit{MaskFusion}~\cite{runz_maskfusion_2018} adds labels to oriented point clouds and supports dense object visualization, but does not maintain persistent objects globally in a graph. \textit{Fusion++}~\cite{fusionPP}, on the other hand, supports persistent dense reconstruction from fixed size $64^3$ voxel grids, yet is sensitive to voxel size tuning and may fail to adapt to objects at varying scales.
Our system utilizes scalable voxel grids that require minimum tuning to adjust object scales. With a seamless CPU to GPU memory transfer implementation, larger environments can be handled on-the-go.

\section{Method} \label{sec: methodology}
Our pipeline can be divided into typical SLAM components and a deep perception module, connected by a object-based semantic map. Figure~\ref{fig:overview} provides an overview.

It consists of 5 modules each running in a separate thread: semantic segmentation, frame-to-model odometry, object data association and map update, PGO, and compositional rendering.
Incoming \textit{RGB-D} frames are initially processed through  \textit{semantic segmentation} (\S\ref{subsec: segmentation})  to obtain instance masks, labels, and semantic descriptors, from DNNs for keyframes.
Then, we estimate the \textit{frame-to-model odometry} (\S\ref{subsec: tracking}) between the incoming live frame and the \textit{compositional render} (\S\ref{subsec: rendering}) from the map to obtain relative poses.
Maintained objects visible in the frame are rendered given the estimated camera pose, and objects are associated with 2D instance detections to either integrate or initialize new objects in the global map (\S\ref{subsec: segmentation}).
Separately, a global factor-graph is updated to optimize the camera trajectory and object poses (\S\ref{subsec: optimization}). The optimized object poses are rendered to generate a compositional model of the scene for subsequent tracking (\S\ref{subsec: rendering}).

Before we discuss these modules in detail from \S\ref{subsec: tracking} to \S\ref{subsec: rendering}, we first introduce core concepts and notations in \S\ref{subsec: notation}.

\subsection{Core concepts and notations} \label{subsec: notation}
A background volume  $V_B$ is a spatially-hashed voxel grid on GPU, where small $16^3$ subvolumes are allocated around observed 3D points. It is created and updated \textit{temporarily} for stable tracking. An object volume $V_{O_i}$ is akin to the background volume, but persistently maintains the object label, ID, and corresponding object descriptors.

A 3D volume $V$'s properties, including surface vertex positions, normals, and colors, can be mapped to 2D images given a camera pose $\mathbf{T} \in SE(3)$ and intrinsics $K$ with ray-casting. We denote such \textit{rendered images} by \( \langle \mathcal{N}, \mathcal{V}, \mathcal{C} \rangle \)  for normal, vertex, and color maps respectively. They can be associated with \textit{input RGB-D images} $\langle \mathcal{I}, \mathcal{D}\rangle $ that consist of color ($\mathcal{I}$) and depth ($\mathcal{D}$) images via projective closest points.

We use subscripts and superscripts to indicate multiple coordinate frames used in our pipeline, including $C_i$ for $i$th camera, $O_j$ for $j$th object, and $W$ for background or world coordinate frame. For instance,
\(\langle \mathcal{N}_{C_s}, \mathcal{V}_{C_s}, \mathcal{C}_{C_s} \rangle \) represents 2D maps rendered from the volumes in the $C_s$ camera coordinate frame.
$\mathbf{T}_{O_j}^W \in SE(3)$ encodes a rigid transformation from object $j$ to world. Finally, we define the camera projection matrix with $K$.



\subsection{Hybrid frame-to-model odometry} \label{subsec: tracking}

In \textit{RGB-D} camera tracking, we seek to estimate the relative camera pose \(\mathbf{T}^{C_t}_{C_s}\) given an incoming \textit{RGB-D} target frame \( \langle \mathcal{I}_{C_t}, \mathcal{D}_{C_t}\rangle \) and a source model \( \langle \mathcal{N}_{C_{s}}, \mathcal{V}_{C_{s}}, \mathcal{C}_{C_{s}}\rangle \) of the scene rendered at the previous camera frame ${C_s}$.

We accomplish this by minimizing the joint weighted dense geometric error residual $r_D$ and the photometric error residual $r_I$. The general energy function is formulated as in \cite{park_colored_2017} by accumulating residual at every point $p \in \mathbb{R}^2$ with a valid data association:
\begin{multline}
    E(\mathbf{T}^{C_t}_{C_{s}}) = \sum_{p} (1 - \sigma) r_I^2(\mathbf{T}^{C_t}_{C_s}, p)
    + \sigma r_D^2(\mathbf{T}^{C_t}_{C_s}, p), \label{eq:1}
\end{multline}
Here, we adapt the geometric ICP residual as the point-to-plane distance between the incoming depth map $\mathcal{D}$ and the rendered vertex and normal map $(\mathcal{V}_{C_s}, \mathcal{N}_{C_s})$ as follows, using the formulation in \cite{kinectfusion}:
\begin{align}
    r_D(T^{C_t}_{C_s}, p) = \bigg((T^{C_t}_{C_{s}} \mathcal{V}_{C_s}( \hat{p} ) - \mathcal{V}_{C_{t}}(p )\bigg) \cdot \mathcal{N}_{C_{t}}( p ), \label{eq:2}
\end{align}
where $\mathcal{V}_{C_t}$ is the vertex map from unprojecting the input depth image $\mathcal{D}_{C_t}$.
Additionally, we use a photometric error residual to improve tracking robustness, which is defined as:
\begin{align}
    r_I(T^{C_t}_{C_s}, p) = \mathcal{C}_{C_s}(\hat{p}) - \mathcal{I}_{C_t}(p). \label{eq:3}
\end{align}
In equations \ref{eq:2} and \ref{eq:3}, $\hat{p}$ is the correspondence of $p$ in the source frame, and is computed via \textit{warping}:
\begin{align}
    \hat{p} = K {\mathbf{T}^{C_t}_{C_s}}^{-1}\mathcal{D}_{C_t}(p)K^{-1}[p^\top, 1]^\top. \label{eq:4}
\end{align}
It must be noted that the $p$s are a subset of pixels with valid object-level data associations detailed in \S\ref{subsec: rendering}.

The energy function in equation \ref{eq:1} is minimized using the Gauss-Newton algorithm. We implement the minimization in a coarse to fine scheme using an image pyramid, on the GPU in parallel since each pixel acts independently in the energy function using \textit{reduction} with appropriate thread conflict handling as described in \cite{dong2019gpu}.

\begin{figure}[t!]
    \centering
    \subfloat{\includegraphics[width=0.5\linewidth]{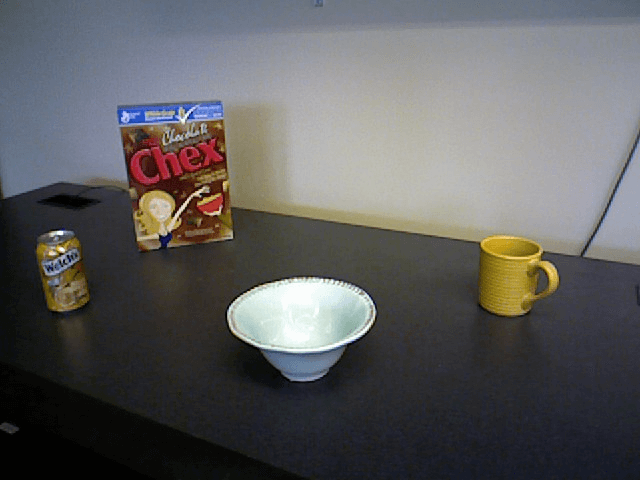}}
    \subfloat{\includegraphics[width=0.5\linewidth]{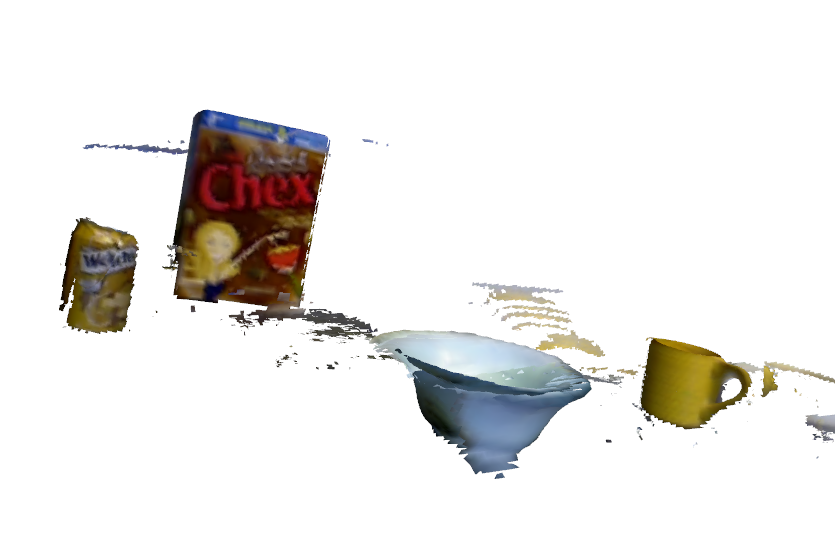}}
    \caption{Qualitative foreground object reconstruction results on \emph{RGB-D Scene 13} sequence.}
    \vspace*{-1em}
    \label{fig:rgbd_scene13}
\end{figure}

\begin{figure*}[t!]
    \centering
    \subfloat{\includegraphics[align=c,width=0.3\linewidth]{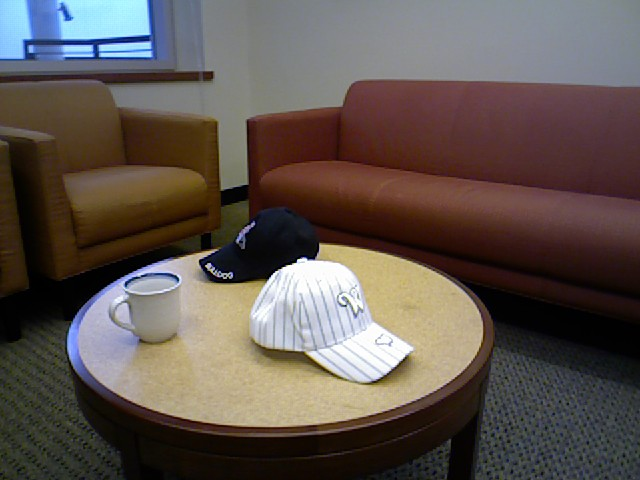}}
    \subfloat{\includegraphics[align=c,width=0.3\linewidth]{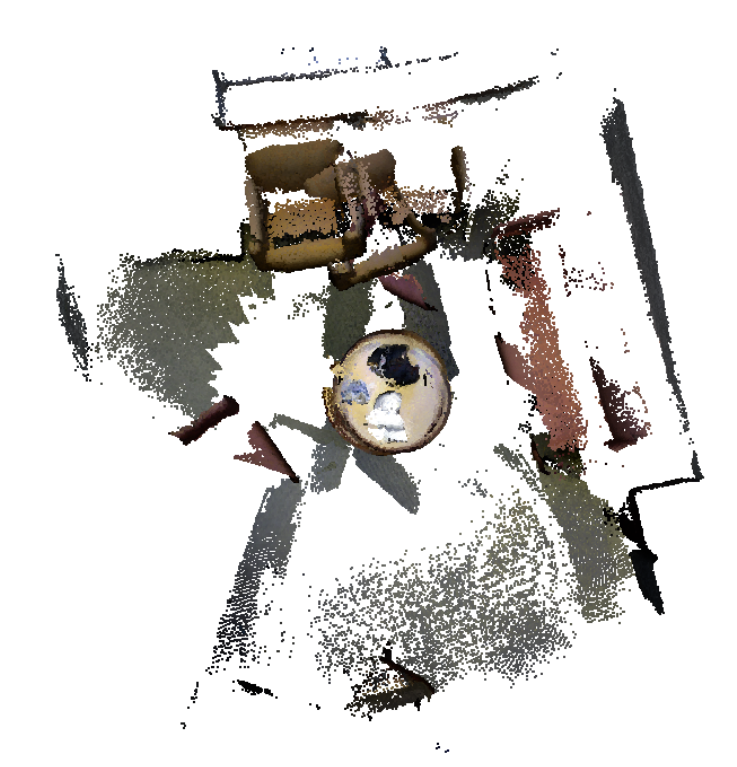}}
    \subfloat{\includegraphics[align=c,width=0.26\linewidth]{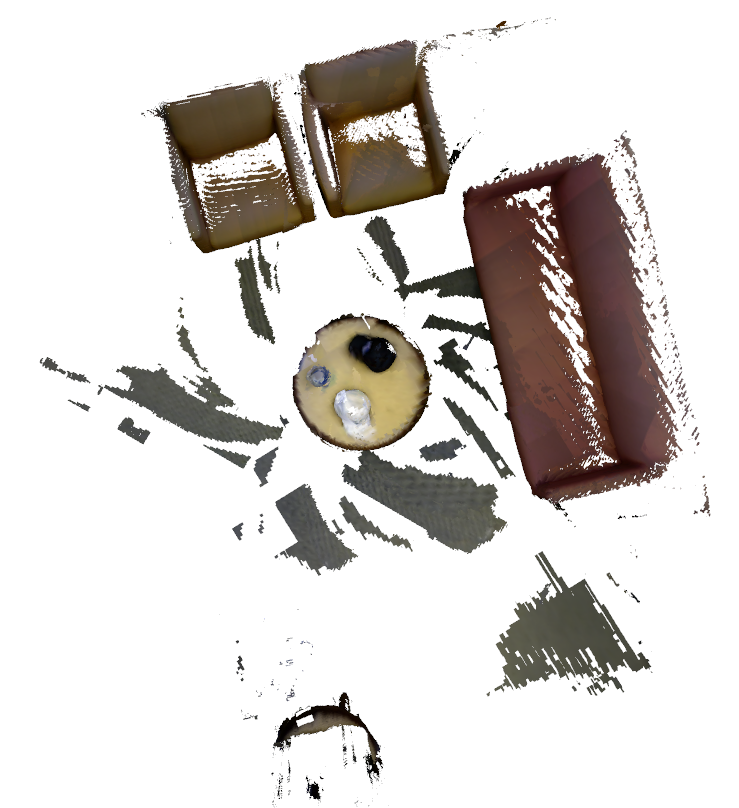}} 
    \vspace{-2mm}
    \caption{Reconstructed small indoor scene \emph{RGB-D Scene 12}. We first show an example input \textit{RGB} frame followed by a top-down view of the reconstruction from \textit{MaskFusion}. This is followed by result from our pipeline. Note that in our reconstruction background walls and floor are filtered out. }
    \vspace*{-1em}
    \label{fig:rgbd_scene12}
\end{figure*}

\subsection{Object instance segmentation and association} \label{subsec: segmentation}

\textbf{2D instance segmentation.} Object detection and instance masks are generated every $n$th frame (we choose $n=10$) in a separate thread from the \textit{PointRend} backend. \textit{PointRend} uses a Resnet-50-FPN backbone network to generate a convolutional feature map. In particular, after an empirical evaluation, we found that \textit{PointRend} provided better masks over \textit{Mask-RCNN}.

The semantic segmentation module maps incoming \textit{RGB} frame \(\mathcal{I}\) into a set of object labels $[l_1, \dots l_k]$, a set of binary object masks $M_n^i$ defined over $l \in \mathcal{L} \triangleq \{0, \dots, L_{max}-1\}$ object classes ($L_{max}=80$ in the MS-COCO dataset), bounding boxes $b \in \mathbb{N}^4$, and a probability distribution $p(l_i \mid \mathcal{I})$. We also extract the object feature map for the accepted object proposals, from the penultimate fully connected layer of the R-CNN from the object classifier head. We observe that these feature maps provide us with a robust data association in ambiguous situations.
To obtain segmentation from frames not sent to the DNN, we warp the binary mask images from the most recent frame with a detection and fill the holes in the masks using the \textit{flood fill} algorithm.

Once the current camera pose and the semantic segmentation information are available, instance detections are associated with existing objects. Unmatched instance detections are used to initialize new object volumes.

\textbf{3D instance generation.} When an object is to be instantiated, the masked depth frame at $C_i$ is unprojected and transformed into the world frame to obtain the object point cloud:
\begin{align}
X_{W} = \mathbf{T}^W_{C_i} K^{-1} D_{C_i}(p) [p^\top, 1]^\top.
\end{align}
To obtain relatively high fidelity reconstruction, we adaptively calculate a conservative voxel length of
\begin{align}
    l = \gamma \| \max(X_W) - \min(X_W) \|_{\infty},
\end{align}
where $\min, \max$ operators are applied to all dimensions of $X \in \mathbb{R}^3$ simultaneously. We empirically use $\gamma = 1 / 64\sqrt{2}$, but due to the scalability of the volume our model is less sensitive to $\gamma$.
Finally, the object pose is simply chained by
\begin{align}
    \mathbf{T}_{O}^W = \mathbf{T}_{C_i}^{W} \bigg({\mathbf{T}_{C_i}^O}\bigg)^{-1},
\end{align}
where $\mathbf{T}_{C_i}^O = [I \mid t_{C_i}^O]$ with $t_{C_i}^O = \min(X_W) - t_{C_i}^W $. Each new object is also initialized with the object feature map from its corresponding instance mask.

\textbf{2D--3D semantic data association}
To associate existing object volumes to 2D instances, visible objects are rendered (described in \ref{subsec: rendering}) in the current frame. The rendered color map $\mathcal{C}$ is thresholded to obtain a virtual binary mask. An intersection over union (IoU) between the virtual binary mask $\hat{\mathcal{M}}$ and the instance masks $\mathcal{M}_i$ in the current frame is used as a scoring metric as defined in \cite{fusionPP}.

As opposed to computing the $\argmax_{i}{\text{IoU}(\mathcal{M}_i, \hat{\mathcal{M})}}$, we associate objects as given below:
\begin{align}
    i = \argmin_{i \in \mathcal{S}} (\| f_i - \hat{f} \|_1),
\end{align}
where $\hat{f}$ and $f_i$ denote feature map of the object render, and the instance masks and $\mathcal{S} \triangleq \{i : \text{IoU}(\mathcal{M}_i, \hat{M}) > 0.2\}$.
Associating object renders to instance masks in this manner prevents incorrectly fusing object instances between nearby similar objects, in cases where there is large accumulated drift.


For subsequent fusion of a 2D instance detection to its associated 3D object, the instance mask---containing the object foreground---and the bounding box mask---containing both the foreground and background are used. Similar to \cite{fusionPP} we integrate the object in both the foreground and background through a weighted average of TSDF, color, and additionally maintain binomial foreground-background count variables for each voxel. This smoothes out artifacts from integration of 2D instances with spurious masks.

Finally, we update the object feature map by a gated weight average:
\begin{align}
    f_t = \frac{w_{t-1} \cdot f_{t-1} + \mathcal{H}(f_{t-1}, f_{in}) \cdot f_{in}}{w_{t-1} + \mathcal{H}(f_{t-1}, f_{in})},\\
    \mathcal{H}(f_{t-1}, f_{in}) = \frac{\textrm{sgn}(\lambda - ||f_{t-1} - f_{in}||_1) + 1}{2},
\end{align}
where $\mathcal{H}$ is the Heaviside step function that hard-filters outlier input feature map $f_{in}$ compared to the maintained object feature map $f_{t-1}$ with weight $w_{t-1}$ controlled by the threshold $\lambda$.

\subsection{Factor graph optimization} \label{subsec: optimization}
As we have mentioned before, a background volume is maintained for stable tracking, and handle \textit{objectless} frames. The background volume, additionally maintains the ratio ($r$) of visible volume units in the current camera frustum to the total number allocated volume units in the volume. A low ratio implies that the camera may have moved away from a particular part of the scene. Pose graph optimization is conditionally triggered when the background volume is reset owing to low ratio of visible units ($r < 0.2$) and when there are new objects added into the graph.

Our object factor graph formulation is similar to \cite{salas-moreno_slam_2013, fusionPP}. The variable nodes $\mathcal{X} = \{\mathbf{x}_1, \dots \mathbf{x}_N\}$ are partitioned into camera pose variables $\mathbf{T}^{W}_{C_i} \in SE(3)$ and object pose variables $\mathbf{T}^W_{O_j} \in SE(3)$. The first camera pose is initialized as the world frame $W$.

Assuming a Gaussian noise model, the \textit{MAP} inference problem with the above variable nodes reduces to solving the following non-linear least squares optimization:
\begin{multline}
    \mathcal{X}^* = \argmin_{\mathcal{X}} \Big( \sum_{k \in |C|} \|  \mathbf{T}^{C_k}_{C_{k-1}} \ominus \mathbf{Z}^{C_k}_{C_{k-1}} \|_{\Sigma_{k,k-1}}^2 \\
    + \sum_{j \in |\mathcal{O}|, k \in |\mathcal{C}|} \| \mathbf{T}^{O_j}_{C_k} \ominus \mathbf{Z}^{O_j}_{C_k} \|_{\Sigma_{o_j, k}}^2 \Big),
\end{multline}
where the operator $ \mathcal{Y} \ominus \mathcal{X} = \text{Log}(\mathcal{X}^{-1} \mathcal{Y})$ expresses the relative error in the local tangent vector space \cite{sola2018micro}. $\Sigma_{k, k-1}$ denotes the covariance between relative camera pose measurements,  $\Sigma_{o_j, k}$ is the covariance in the camera to object measurement. They can be approximated by information matrices computed from \textit{odometry}, however, empirically we found that a constant information matrix can achieve reasonable results. We obtain the relative camera measurements from the \textit{frame to model odometry} (\S\ref{subsec: tracking}), and perform an additional Gauss Newton iteration with only the object pixels, to obtain frame to object measurements. Finally, the expected relative camera pose $\mathbf{T}^{C_k}_{C_{k-1}}$ and expected camera object pose $\mathbf{T}^{O_j}_{C_k}$ used in the factors are calculated as:
\begin{align}
    \mathbf{T}^{C_k}_{C_{k-1}} &= \bigg({\mathbf{T}^{W}_{C_k}}\bigg)^{-1} \mathbf{T}^{W}_{C_{k-1}}, \\
    \mathbf{T}^{O_j}_{C_k} &= \bigg({\mathbf{T}^{W}_{O_j}}\bigg)^{-1} \mathbf{T}^{W}_{C_k}.
\end{align}

We solve the optimization in GTSAM \cite{dellaert2012factor}, using Levenberg Marquardt. Object volumes remain unchanged in memory after optimization, since the entire object volume is transformed. We note that this circumvents the time-consuming re-integration that usually takes place in volumetric methods after PGO \cite{whelan2016elasticfusion}.

\subsection{Compositional rendering} \label{subsec: rendering}
\begin{figure}[t!]
    \centering
    \includegraphics[width=\linewidth]{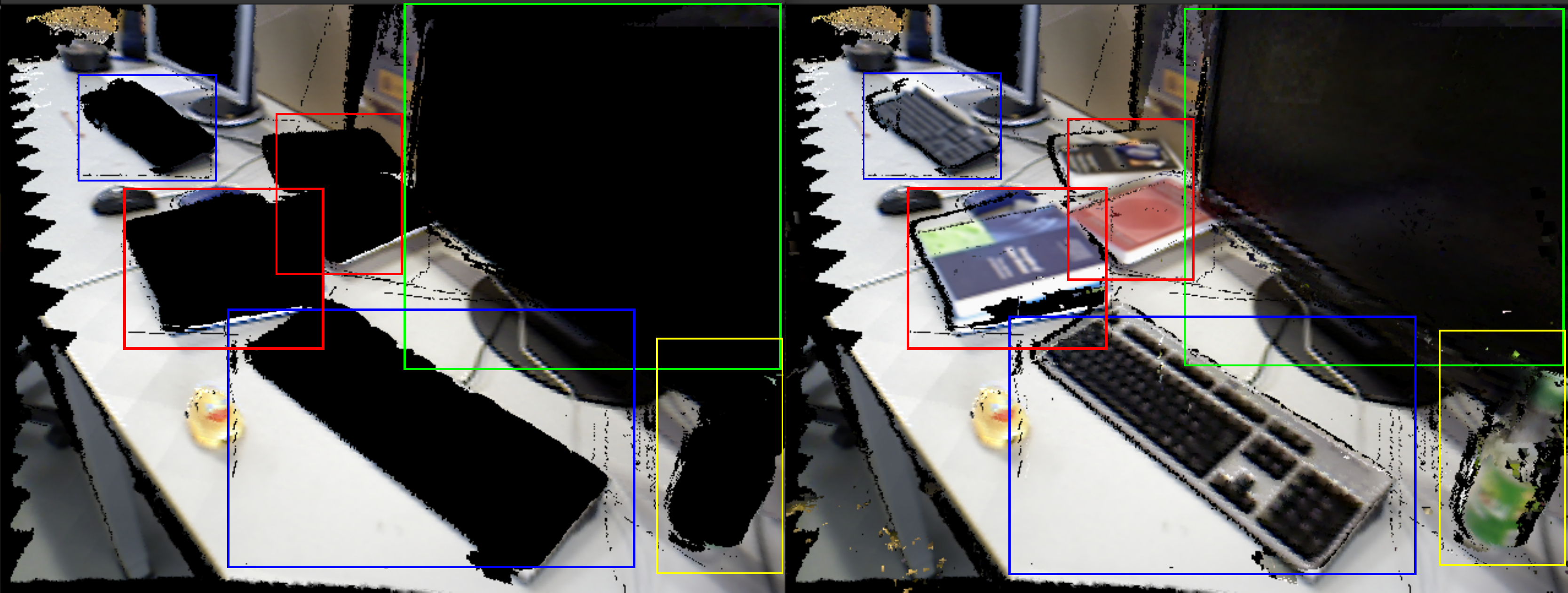}\vspace{-5mm}
    \caption{Compositional rendering during reconstruction on TUM \textit{fr3\_long\_office\_household} dataset. Left shows the background render, and right shows the composed render. Compositionally rendered objects are shown in bounding boxes.}
    \vspace*{-1em}
    \label{fig:compositional_render}
\end{figure}

Compositional rendering is a serialized operation that generates normal, vertex, and color maps by ray-casting 3D objects in the viewing frustum into 2D object instances, and is illustrated in Figure~\ref{fig:compositional_render}.

\(\langle \mathcal{N}_{C_i}, \mathcal{V}_{C_i}, \mathcal{C}_{C_i} \rangle \) is in fact an aggregation of separate renderings from object volumes \(\langle \mathcal{N}_{C_i}^{V_{{O}_j}}, \mathcal{V}_{C_i}^{V_{{O}_j}}, \mathcal{C}_{C_i}^{V_{\mathcal{O}_j}} \rangle \) and the background volume \(\langle \mathcal{N}_{C_i}^{V_{B}}, \mathcal{V}_{C_i}^{V_{B}}, \mathcal{C}_{C_i}^{V_{{B}}}  \rangle \), depending on the masks. In particular, we render the background volume, based on a background mask that is constructed as a union of existing virtual object masks in the current frame, and associated instance masks.

Then, the composed per-pixel map model render can be obtained as follows:
\begin{align}
    &\hat{k} = \argmin_k \mathcal{V}_k(p)[z], ~k \in \{O_1, \cdots, O_n, B\} \\
    &\langle \mathcal{N}^*(p), \mathcal{V}^*(p), \mathcal{C}^*(p) \rangle = \langle  \mathcal{N}_{\hat{k}}(p), \mathcal{V}_{\hat{k}}(p), \mathcal{C}_{\hat{k}}(p) \rangle,
\end{align}
where $\hat{k}$ is the volume index corresponding to the minimum distance to camera center for pixel $p$.

Object volumes not currently visible are downloaded from GPU into CPU memory. Note that downloading the object volume does not affect the optimization problem, since the object volumes are required only for integration and raycasting.


\begin{table*}[htbp]
\centering
\caption{Trajectory accuracy comparison on Realworld dataset (Absolute Trajectory Error in centimeters)}
\resizebox{0.82\linewidth}{!}{
\begin{tabular}{ |p{3cm}||p{1.5cm}|p{1.5cm}||p{1.5cm}| p{1.5cm} | p{1.5cm}|  }
 \hline
 Dataset & ElasticFusion & Open3D & MaskFusion & Fusion++ & Ours \\
 \hline
 Online & \checkmark &  & \checkmark & \checkmark & \checkmark \\
 \hline
 Object Models & & & \checkmark & \checkmark & \checkmark \\
 \hline \hline
RGBD Scenes - Scene 03 & 1.42 & 19.37 & 26.67 & - & \textbf{4.52}\\ 
\hline
RGBD Scenes - Scene 12 & 0.64 & 1.97 & 10.81 & - & \textbf{2.36}\\
\hline
RGBD Scenes - Scene 14 & 1.09 & 1.33 & 8.26 & - & \textbf{2.37}\\
\hline
freiburg1\_xyz & 6.33 & 6.64 & 8.68 & - & \textbf{7.50}\\
\hline
freiburg1\_desk & 2.70 & 5.73 & 24.05 & \textbf{4.9} & 5.82\\
\hline
freiburg1\_desk2 & 7.12 & 7.65 & 21.5 & 15.3 & \textbf{9.57}\\
\hline
freiburg1\_room & 22.06 & 5.65 & 52.4 & 23.5 & \textbf{21.7}\\
\hline
freiburg2\_xyz & 1.12 & 2.18 & 12.30 & \textbf{2.0} & 2.27\\
\hline
freiburg2\_desk & 7.61 & 4.72 & 163.6 & 11.4 & \textbf{9.94}\\
\hline
freiburg3\_long\_office & 2.23 & 3.54 & 140.8 & 10.8 & \textbf{9.68}\\
\hline
\end{tabular}
}
\label{tab:ATE_RMSE}
\end{table*}

\section{System Architecture}

To support relatively high frame rate operation in the presence of slow/non-realtime deep learning components our pipeline is highly parallelized. Our system adopts the \textit{Actor} framework, where each component runs asynchronously, and communicates via thread-safe queues.

We implement the semantic segmentation pipeline as a separate python process which serializes the outputs using \texttt{protobuf} and communicates with the client thread via \texttt{zeromq} sockets in the Object SLAM pipeline. Since, instance segmentation is carried out only for keyframes, typically, the asynchronous python process exits early freeing GPU memory for larger scene reconstructions.

The GPU code is implemented in CUDA, and we leverage the Open3D framework \cite{dong2019gpu} for a robust scalable TSDF implementation.

\section{Experimental Results} \label{sec: results}

In this section, we show that our system achieves comparable results to state-of-the-art online/offline reconstruction systems in terms of trajectory accuracy while being able to segment and reconstruct objects.  We evaluate on the RGBD scenes V2 dataset \cite{lai_unsupervised_2014} and TUM RGBD dataset \cite{sturm12iros}, both of which are established RGBD benchmarks and compare against baselines. Our experiments were run on a Linux system with Intel i7-6700 CPU at 4.00GHz and 32GB of RAM and a NVIDIA GTX1080 with 8GB of GPU memory.

\subsection{Qualitative results}

We first demonstrate qualitative reconstruction results on the RGBD scenes V2 dataset.  Fig.~\ref{fig:rgbd_scene13} shows the object mesh extracted from our scalable volumes with a foreground count threshold. We can see that small objects are clearly reconstructed with details, and the background is correctly filtered. 

At a larger scale, Fig.~\ref{fig:objectsandscene} segments teddy bear and computers from the cluttered scene and ensures a low-drift of the trajectory. Fig.~\ref{fig:rgbd_scene12} compares reconstructions from \textit{MaskFusion} \cite{runz_maskfusion_2018} and our system for a given sequence. It can be seen that the object-level reconstruction, specifically for caps and sofas, is much cleaner by our system than by \textit{MaskFusion}. 


\subsection{Quantitative results}

Table \ref{tab:ATE_RMSE} presents Absolute Trajectory Error (ATE) of four different methods compared with our system. Note in the table we \textit{bold} the best results of \textit{the object-based systems}, and provide results from geometric SLAM systems for reference.

In general, we achieve comparable results against the state-of-the-art surfel based online SLAM system \textit{ElasticFusion}~\cite{whelan2016elasticfusion} and volumetric offline reconstruction system \textit{Open3D}~\cite{zhou_open3d_2018}. In the meantime, our method outperforms object-based SLAM systems \textit{MaskFusion}\footnote{MaskFusion can only work in online mode with 2 high end graphics cards. We ran in offline mode, and kept all detected object instances instead of manually selecting objects of interest for a fair comparison.}~\cite{runz_maskfusion_2018}
and \textit{Fusion++} \footnote{Fusion++ is not open sourced and we obtained the results from the paper.}\cite{fusionPP} by a large margin. 

\begin{figure}[t!]
    \centering
    \subfloat[\label{sfig:obj_loop_closure_a}]{\includegraphics[width=0.5\linewidth]{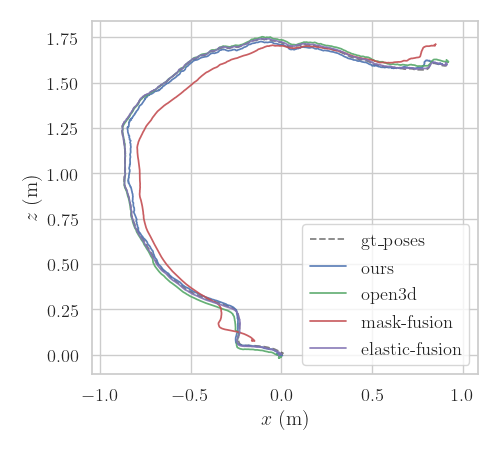}}
    \subfloat[\label{sfig:obj_loop_closure_b}]{\includegraphics[width=0.5\linewidth]{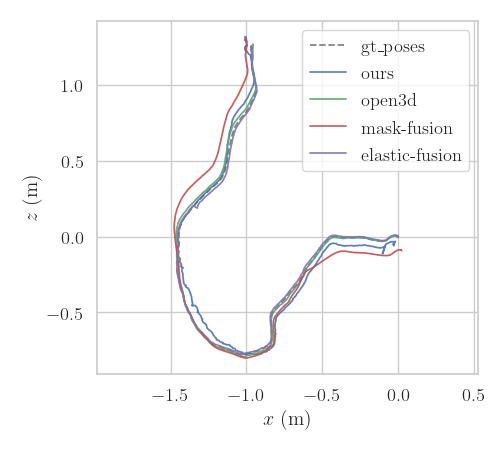}}
    \caption{Comparison of trajectories between our pipeline and baselines with ground truth. (a) shows \emph{rgbd-scenes-v12} and (b) shows \emph{rgbd-scenes-v14} sequences. }
    \vspace*{-1em}
    \label{fig:rgbd_scenes-ape}
\end{figure}

For small scenes in the \textit{RGB-D scenes V2 dataset}, we achieve consistently high accuracy with ATE below $5cm$ for all scenes. Figure~\ref{fig:rgbd_scenes-ape} shows detailed trajectory visualizations.

\begin{figure}[t!]
    \centering
    \subfloat[]{\includegraphics[width=0.5\linewidth]{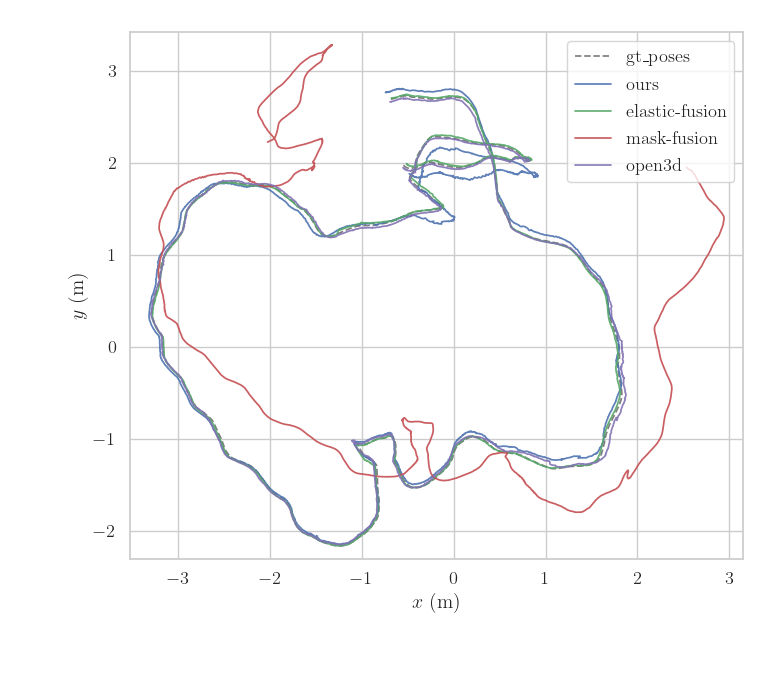}}
    \subfloat[]{\includegraphics[width=0.5\linewidth]{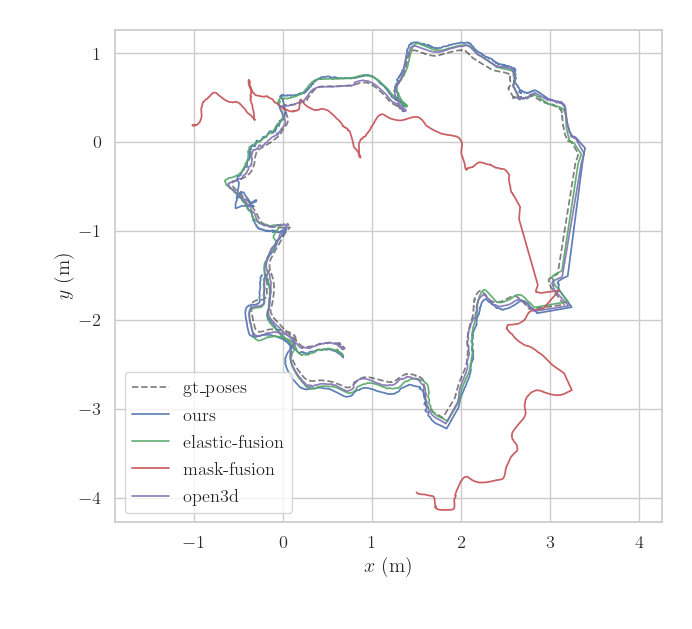}}
    \caption{Trajectory comparisons on the TUM-RGBD dataset (a) \emph{fr3\_long\_office\_household} and (b) \emph{fr2\_desk} sequences showing that even in the absence of explicit loop closures, our system maintains comparable accuracy.}
    \vspace*{-1em}
    \label{fig:fr3_household-ape}
    \vspace{12mm}
\end{figure}

For larger scenes, although noisy semantic segmentations affect masks and introduce noise for frame-to-model odometry, compositional rendering still ensures reliable tracking. Trajectory comparisons are provided in Figure~\ref{fig:fr3_household-ape}.


\subsection{Runtime analysis}

We evaluated the runtime of our system on the above mentioned scenes by limiting the number of objects initialized to 10. Processing each frame in the absence of any objects i.e., only background tracking takes about 200ms per frame. The largest computational bottleneck and time consuming operation is the rendering step, and while there are multiple rendering operations required (for instance, during object association), we render the objects and background only once per frame, and reuse the renders. We observe that each object takes on average about 45ms to render. In the presence of about 5-8 objects in the scene, the time taken per frame increases to about 450ms per frame. In comparison, we observed in our tests that \textit{MaskFusion} runs at lower than 1FPS with one graphics card and suffers from random crashes in online mode. \textit{Fusion++} reports its results with pre-computed segmentation masks. Our method runs seamlessly on a single GPU. A detailed runtime analysis is given in Table~\ref{tab:runtime}.

\begin{table}[ht]
    \centering
    \caption{Runtime breakdown component-wise for our pipeline}    
    \resizebox{0.995\linewidth}{!}{
    \begin{tabular}{|c|c|c|c|c|c}
        \hline
        Component & Tracking & Segmentation & Association & Rendering\\
        \hline
         Time (ms) & 13 & 250 & 15 & 45 per object\\
        \hline
    \end{tabular}
    }
    \label{tab:runtime}
\end{table}

\section{Conclusions and Future Work}
In this paper, we have presented Compositional and Scalable Object SLAM, which bridges geometry-based techniques effectively with deep object detection. The system maintains persistent independent 3D models of the objects visible in the scene, which provides for relatively accurate trajectories as well as object reconstructions.

While our system makes significant progress towards semantic SLAM, it is not perfect owing to the following shortcomings. 1) TSDF inpainting is not considered, causing partially reconstructed objects; the system slows down subsequently with increasing map size, albeit at a slow rate. 2) Object labels are limited to 80 classes from the MS-COCO dataset; 3) Finally in the presence of temporal instance switches and missed detections of small objects, tracking accuracy is affected. 

Solving these shortcomings provide avenues for interesting future work. In particular, we plan to introduce object model based reconstructions and neural rendering for better object reconstructions.


\balance

\bibliographystyle{IEEEtran}
\bibliography{references,ref}

\end{document}